\documentclass[twoside]{article}
\usepackage[accepted]{aistats2017}

\usepackage{algorithm}
\usepackage[noend]{algorithmic}

\newcommand{\MyAnd}{ \hspace{5mm} }
\newcommand{\RunSpc}{ \hspace{0.5mm} }

\usepackage{hyperref}

\usepackage{booktabs}       %
\usepackage{amsfonts}       %
\usepackage{nicefrac}       %
\usepackage{microtype}      %
\usepackage{subcaption}

\usepackage[numbers]{natbib}
\usepackage{enumitem}
\setlist{nolistsep}
\setlist[itemize]{noitemsep, topsep=0pt}

\usepackage{times}

\usepackage{graphicx} %
\usepackage{color}

\usepackage{array}
\usepackage{amssymb}
\usepackage{amsmath}
\usepackage{xspace}
\usepackage{fancyhdr}
\usepackage{comment}

\newcommand{\thetitle}{Communication-Efficient Learning of Deep Networks \\
from Decentralized Data}
\newcommand{\titlenobreak}{Communication-Efficient Learning of Deep Networks
from Decentralized Data}

\newcommand{\mycaptionof}[2]{\captionof{#1}{#2}}

\newcolumntype{H}{>{\setbox0=\hbox\bgroup}c<{\egroup}@{}}

\newcommand{\noaistats}[1]{}  %

\definecolor{darkgreen}{rgb}{0,0.4,0.0}
\definecolor{darkblue}{rgb}{0,0.1,0.3}
\definecolor{darkred}{rgb}{0.7,0.0,0.0}

\newcommand{\lbs}{\ensuremath{B}}      %
\newcommand{\all}{\infty}
\newcommand{\lepochs}{\ensuremath{E}}  %
\newcommand{\clientfrac}{\ensuremath{C}}    %
\newcommand{\tnn}{2NN\xspace} %
\newcommand{\xx}{\hspace{-0.01in}\ensuremath{\times}}
\newcommand{\loss}{\ell}
\newcommand{\SUB}[1]{\ENSURE \hspace{-0.15in} \textbf{#1}}
\newcommand{\algfont}[1]{\texttt{#1}}

\newcommand{\fedavglong}{\algfont{FederatedAveraging}\xspace}
\newcommand{\fedavg}{\algfont{FedAvg}\xspace}
\newcommand{\fedavgshort}{\algfont{FedAvg}\xspace}

\newcommand{\fedsgdlong}{\algfont{FederatedSGD}\xspace}
\newcommand{\fedsgd}{\algfont{FedSGD}\xspace}
\newcommand{\fedsgdshort}{\algfont{FedSGD}\xspace}

\newcommand{\eqdef}{\overset{\text{def}}{=}}
\newcommand{\T}{\rule{0pt}{2.2ex}}
\newcommand{\targetTNN}{97\%\xspace}
\newcommand{\targetCNN}{99\%\xspace}
\newcommand{\targetLSTM}{54\%\xspace}
\newcommand{\nc}{K}
\newcommand{\pp}{\mathcal{P}}

\newlength{\pw}

\newcommand{\BO}{\mathcal{O}}
\newcommand{\R}{\ensuremath{\mathbb{R}}}

\newcommand{\qqwhere}{\qquad \text{where} \qquad}

\DeclareMathOperator*{\E}{\mathbb{E}}
\newcommand{\grad}{\triangledown}
\newcommand{\h}{\frac{1}{2}}

\begin{document}

\runningtitle{\titlenobreak}

\runningauthor{%
H. Brendan McMahan, \RunSpc
Eider Moore, \RunSpc
Daniel Ramage, \RunSpc
Seth Hampson, \RunSpc
Blaise Ag\"{u}era y Arcas}

\twocolumn[
\aistatstitle{\thetitle}

\aistatsauthor{%
  H. Brendan McMahan \MyAnd
  Eider Moore \MyAnd
  Daniel Ramage \MyAnd
  Seth Hampson \MyAnd
  Blaise Ag\"{u}era y Arcas}
\aistatsaddress{Google, Inc., 651 N 34th St., Seattle, WA 98103 USA}
]

\begin{abstract}
Modern mobile devices have access to a wealth of data suitable for learning models, which in turn can greatly improve the user experience on the device. For example, language models can improve speech recognition and text entry, and image models can automatically select good photos. However, this rich data is often privacy sensitive, large in quantity, or both, which may preclude logging to the data center and training there using conventional approaches.  We advocate an alternative that leaves the training data distributed on the mobile devices, and learns a shared model by aggregating locally-computed updates. We term this decentralized approach \emph{Federated Learning}.

We present a practical method for the federated learning of deep networks based on iterative model averaging, and conduct an extensive empirical evaluation, considering five different model architectures and four datasets. These experiments demonstrate the approach is robust to the unbalanced and non-IID data distributions that are a defining characteristic of this setting. Communication costs are the principal constraint, and we show a reduction in required communication rounds by 10--100$\xx$ as compared to synchronized stochastic gradient descent. 
\end{abstract}

\section{Introduction}
\label{intro}

Increasingly, phones and tablets are the primary computing devices for
many people \citep{pew16smartphone,pew15deviceownership}. The powerful sensors
on these devices (including cameras, microphones, and GPS), combined
with the fact they are frequently carried, means they have
access to an unprecedented amount of data, much of it private in
nature. Models learned on such data hold the promise of greatly
improving usability by powering more intelligent applications,
but the sensitive nature of the data means there
are risks and responsibilities to storing it in a centralized
location.

We investigate a learning technique that allows users to collectively
reap the benefits of shared models trained from this rich data,
without the need to centrally store it.  \noaistats{This approach also allows us
to scale up learning by utilizing the plentiful computation available at
the edges of the network, conveniently colocated with the data.}
We term our approach \emph{Federated Learning}, since the learning
task is solved by a loose federation of participating devices (which
we refer to as \emph{clients}) which are coordinated by a central
\emph{server}. Each client has a local training dataset which is never
uploaded to the server. Instead, each client computes an update to the
current global model maintained by the server, and only this update is
communicated.  This is a direct application of the principle of
\emph{focused collection} or \emph{data minimization} proposed by the
2012 White House report on privacy of consumer
data~\citep{whitehouse13privacy}.  Since these updates are specific to
improving the current model, there is no reason to store them once
they have been applied.

A principal advantage of this approach is the decoupling of model training from the need for direct access to the raw training data. Clearly, some trust of the server coordinating the training is still required. However, for applications where the training objective can be specified on the basis of data available on each client, federated learning can significantly reduce privacy and security risks by limiting the attack surface to only the device, rather than the device and the cloud.

Our primary contributions are 1) the identification of the problem of
training on decentralized data from mobile devices as an important
research direction; 2) the selection of a straightforward and
practical algorithm that can be applied to this setting; and 3) an
extensive empirical evaluation of the proposed approach. More
concretely, we introduce the \fedavglong algorithm, which combines
local stochastic gradient descent (SGD) on each client with a server
that performs model averaging. We perform extensive experiments on
this algorithm, demonstrating it is robust to unbalanced and non-IID
data distributions, and can reduce the rounds of communication needed
to train a deep network on decentralized data by orders of magnitude.

\paragraph{Federated Learning}

Ideal problems for federated learning have the following properties:
1) Training on real-world data from mobile devices provides a distinct
advantage over training on proxy data that is generally available in
the data center. 2) This data is privacy sensitive or large in size
(compared to the size of the model), so it is preferable not to log it
to the data center purely for the purpose of model training (in
service of the \textit{focused collection} principle).
3) For supervised tasks, labels on the data can be inferred naturally
from user interaction.

Many models that power intelligent behavior on mobile devices fit the
above criteria. As two examples, we consider \emph{image
  classification}, for example predicting which photos are most likely
to be viewed multiple times in the future, or shared; and \emph{language
  models}, which can be used to improve voice recognition and text
entry on touch-screen keyboards by improving decoding,
next-word-prediction, and even predicting whole
replies~\citep{corrado15computer}.
The potential training data for both these tasks (all the photos a
user takes and everything they type on their mobile keyboard,
including passwords, URLs, messages, etc.) can be privacy sensitive.
The distributions from which these examples are drawn are also likely
to differ substantially from easily available proxy datasets: the use
of language in chat and text messages is generally much different than
standard language corpora, e.g., Wikipedia and other web documents; the photos people take on their phone are likely
quite different than typical Flickr photos. And finally, the labels
for these problems are directly available: entered text is
self-labeled for learning a language model, and photo labels can be
defined by natural user interaction with their photo app (which photos
are deleted, shared, or viewed).

Both of these tasks are well-suited to learning a neural network. For
image classification feed-forward deep networks, and in
particular convolutional networks, are well-known to provide
state-of-the-art results
\citep{lecun98gradientbased,krizhevsky12imagenet}. For language
modeling tasks recurrent neural networks, and in particular LSTMs,
have achieved state-of-the-art results
\citep{hochreiter97lstm,bengio03neural,kim15character}.

\paragraph{Privacy}

Federated learning has distinct privacy advantages compared to data
center training on persisted data. Holding even an ``anonymized''
dataset can still put user privacy at risk via joins with other
data~\citep{sweeney00simple}. In contrast, the information transmitted
for federated learning is the minimal update necessary to improve a
particular model (naturally, the strength of the privacy benefit
depends on the content of the updates).\footnote{ For example, if the
  update is the total gradient of the loss on all of the local data,
  and the features are a sparse bag-of-words, then the non-zero
  gradients reveal exactly which words the user has entered on the
  device. In contrast, the sum of many gradients for a dense model
  such as a CNN offers a harder target for attackers seeking
  information about individual training instances (though attacks are
  still possible).} The updates themselves can (and should) be
ephemeral. They will never contain more information than the raw
training data (by the data processing inequality), and will generally
contain much less. Further, the source of the updates is not needed by
the aggregation algorithm, so updates can be transmitted without
identifying meta-data over a mix network such as Tor
\citep{chaum81untraceable} or via a trusted third party.
We briefly discuss the possibility of combining federated learning
with secure multiparty computation and differential privacy at the
end of the paper.

\noaistats{\paragraph{Advantages for large datasets}
Federated learning can also provide a distinct advantage when training
on large volumes of data. The network traffic per-client necessary to
train in the data center is simply the size of a client's local
dataset, which must be transmitted once; for federated learning, the
per-client traffic is
$(\text{\#-communication-rounds})\times(\text{update-size})$. This
latter quantity can be substantially smaller if the update-size
(generally $\BO(\text{\#-model-parameters})$) is relatively small
compared to the volume of training data needed, as when training on high-resolution photos or videos.}

\paragraph{Federated Optimization} %

We refer to the optimization problem implicit in federated learning as
federated optimization, drawing a connection (and contrast) to
distributed optimization. 
Federated optimization has several key properties
that differentiate it from a typical distributed optimization
problem:
\begin{itemize}
\item \textbf{Non-IID} The training data on a given client is
  typically based on the usage of the mobile device by a particular
  user, and hence any particular user's local dataset will not be
  representative of the population distribution.
\item \textbf{Unbalanced} Similarly, some users will make much heavier
  use of the service or app than others, leading to varying amounts of
  local training data.
\item \textbf{Massively distributed} We expect the number of clients
  participating in an optimization to be much larger than the average
  number of examples per client.
\item \textbf{Limited communication} Mobile devices are frequently
  offline or on slow or expensive connections.
\end{itemize}
In this work, our emphasis is on the non-IID and unbalanced properties
of the optimization, as well as the critical nature of the
communication constraints.
A deployed federated optimization system must also address a myriad of
practical issues: client datasets that change as data is added and
deleted; client availability that correlates with the local data
distribution in complex ways (e.g., phones from speakers of
  American English will likely be plugged in at different times than
  speakers of British English); and clients that never respond or
send corrupted updates.

These issues are beyond the scope of the current work; instead, we use
a controlled environment that is suitable for experiments, but still
addresses the key issues of client availability and unbalanced and
non-IID data. We assume a synchronous update scheme that proceeds in
rounds of communication.  There is a fixed set of $\nc$ clients, each
with a fixed local dataset. At the beginning of each round, a random
fraction $\clientfrac$ of clients is selected, and the server sends
the current global algorithm state to each of these clients (e.g., the
current model parameters). We only select a fraction of clients for
efficiency, as our experiments show diminishing returns for adding
more clients beyond a certain point. Each selected client then
performs local computation based on the global state and its local
dataset, and sends an update to the server. The server then applies
these updates to its global state, and the process repeats.

While we focus on non-convex neural network objectives, the algorithm
we consider is applicable to any finite-sum objective of the form
\begin{equation}
\label{eq:problem}
\min_{w \in \R^d} f(w)
\qqwhere
f(w) \eqdef \frac{1}{n} \sum_{i=1}^n f_i(w).
\end{equation}
For a machine learning problem, we typically take $f_i(w) = \ell(x_i,
y_i; w)$, that is, the loss of the prediction on example $(x_i, y_i)$
made with model parameters $w$.
We assume there are $\nc$ clients over which the data is partitioned,
with $\pp_k$ the set of indexes of data points on client $k$, with $n_k
= |\pp_k|$.  Thus, we can re-write the objective \eqref{eq:problem} as
\begin{equation*}
\label{eq:problem:distributed}
f(w) = \sum_{k = 1}^K \frac{n_k}{n} F_k(w)
\quad \text{where} \quad
F_k(w) = \frac{1}{n_k} \sum_{i \in \mathcal{P}_k} f_i(w).
\end{equation*}
If the partition $\pp_k$ was formed by distributing the training
examples over the clients uniformly at random, then we would have
$\E_{\pp_k}[F_k(w)] = f(w)$, where the expectation is over the set of
examples assigned to a fixed client $k$. This is the IID assumption
typically made by distributed optimization algorithms; we refer to the
case where this does not hold (that is, $F_k$ could be an arbitrarily
bad approximation to $f$) as the non-IID setting.

In data center optimization, communication costs are relatively small,
and computational costs dominate, with much of the recent emphasis
being on using GPUs to lower these costs. In contrast, in federated
optimization communication costs dominate --- we will typically be
limited by an upload bandwidth of 1 MB/s or less. Further, clients
will typically only volunteer to participate in the optimization when
they are charged, plugged-in, and on an unmetered wi-fi
connection. Further, we expect each client will only participate in a
small number of update rounds per day.
On the other hand, since any single on-device dataset is small
compared to the total dataset size, and modern smartphones have
relatively fast processors (including GPUs), computation becomes
essentially free compared to communication costs for many model
types. Thus, our goal is to use additional computation in order to
decrease the number of rounds of communication needed to train a
model. There are two primary ways we can add computation:
1) \emph{increased parallelism}, where we use more clients working
  independently between each communication round; and, 2)
\emph{increased computation on each client}, where rather than
  performing a simple computation like a gradient calculation, each
  client performs a more complex calculation between each
  communication round.
We investigate both of these approaches, but the speedups we achieve
are due primarily to adding more computation on each client, once a
minimum level of parallelism over clients is used.

\paragraph{Related Work} 
Distributed training by iteratively averaging locally trained models
has been studied
by~\citet{mcdonald10distributed} for the perceptron and
\citet{povey15parallel} for speech recognition DNNs.
\citet{zhang15elastic} studies an asynchronous approach with ``soft''
averaging.  These works only consider the cluster / data center setting
(at most 16 workers, wall-clock time based on fast networks), and do
not consider datasets that are unbalanced and non-IID, properties that
are essential to the federated learning setting.
We adapt this style of algorithm to the federated setting and perform
the appropriate empirical evaluation, which asks different questions
than those relevant in the data center setting, and requires different
methodology.

Using similar motivation to ours, \citet{neverova16motion} also
discusses the advantages of keeping sensitive user data on device. The
work of \citet{shokri15privacy} is related in several ways: they focus
on training deep networks, emphasize the importance of privacy, and
address communication costs by only sharing a subset of the parameters
during each round of communication; however, they also do not consider
unbalanced and non-IID data, and the empirical evaluation is limited.

In the convex setting, the problem of distributed optimization and
estimation has received significant attention
\citep{balcan12distributed, fercoq14fast, shamir14distributed}, and
some algorithms do focus specifically on communication efficiency
\citep{zhang13information,shamir13dane,tianbao13trading,
  chenxin15cocoa, zhang15disco}. In addition to assuming convexity,
this existing work generally requires that the number of clients is
much smaller than the number of examples per client, that the data is
distributed across the clients in IID fashion, and that each node has
an identical number of data points --- all of these assumptions are
violated in the federated optimization setting.  Asynchronous
distributed forms of SGD have also been applied to training neural
networks, e.g., \citet{dean12large}, but these approaches require a
prohibitive number of updates in the federated setting. 
Distributed consensus algorithms (e.g., \citep{zhang14admm}) relax the
IID assumption, but are still not a good fit for
communication-constrained optimization over very many clients.

One endpoint of the (parameterized) algorithm family we consider is
simple one-shot averaging, where each client solves for the model that
minimizes (possibly regularized) loss on their local data, and these
models are averaged to produce the final global model. This approach
has been studied extensively in the convex case with IID data, and it
is known that in the worst-case, the global model produced is no
better than training a model on a single client
\citep{zhang12communication,arjevani15commcomplexity,zinkevich10parallelized}.

\section{The \fedavglong Algorithm}
The recent multitude of successful applications of deep learning have
almost exclusively relied on variants of stochastic gradient descent
(SGD) for optimization; in fact, many advances can be understood as
adapting the structure of the model (and hence the loss function) to
be more amenable to optimization by simple gradient-based methods
\citep{goodfellow16deeplearning}. Thus, it is natural that we build
algorithms for federated optimization by starting from SGD.

SGD can be applied naively to the federated optimization problem,
where a single batch gradient calculation (say on a randomly selected
client) is done per round of communication. This approach is
computationally efficient, but requires very large numbers of rounds
of training to produce good models (e.g., even using an advanced
approach like batch normalization, \citet{ioffe16batchnorm} trained
MNIST for 50000 steps on minibatches of size 60). We consider this
baseline in our CIFAR-10 experiments.

In the federated setting, there is little cost in wall-clock time to
involving more clients, and so for our baseline we use large-batch
synchronous SGD; experiments by \citet{chen16revisiting} show this
approach is state-of-the-art in the data center setting, where it
outperforms asynchronous approaches.  To apply this approach in the
federated setting, we select a $\clientfrac$-fraction of clients on
each round, and compute the gradient of the loss over all the data
held by these clients. Thus, $\clientfrac$ controls the \emph{global}
batch size, with $\clientfrac=1$ corresponding to full-batch
(non-stochastic) gradient descent.\footnote{ While the batch selection
  mechanism is different than selecting a batch by choosing individual
  examples uniformly at random, the batch gradients $g$ computed by
  \fedsgd still satisfy $\E[g] = \grad f(w)$.}  We refer to this
baseline algorithm as \fedsgdlong (or \fedsgdshort).

A typical implementation of \fedsgdshort with $\clientfrac=1$ and a
fixed learning rate $\eta$ has each client $k$ compute $g_k = \grad
F_k(w_t)$, the average gradient on its local data at the current model
$w_t$, and the central server aggregates these gradients and applies
the update $ w_{t+1} \leftarrow w_t - \eta \sum_{k=1}^K \frac{n_k}{n}
g_k, $ since $\sum_{k=1}^K \frac{n_k}{n} g_k = \grad f(w_t)$.
An equivalent update is given by
$\forall k,\ w_{t+1}^k \leftarrow w_t - \eta g_k$
and then
$w_{t+1}   \leftarrow \sum_{k=1}^K \frac{n_k}{n} w_{t+1}^k$.
That is, each client locally takes one step of gradient descent on the
current model using its local data, and the server then takes a
weighted average of the resulting models. 
Once the algorithm is written this way, we
can add more computation to each client by iterating the local update
$w^k \leftarrow w^k - \eta \grad F_k(w^k)$ multiple times before the
averaging step. We term this approach \fedavglong (or \fedavgshort). The
amount of computation is controlled by three key parameters:
$\clientfrac$, the fraction of clients that perform computation on
each round; $\lepochs$, then number of training passes each client
makes over its local dataset on each round; and $\lbs$, the local
minibatch size used for the client updates. We write $\lbs=\all$ to
indicate that the full local dataset is treated as a single minibatch.
Thus, at one endpoint of this algorithm family, we can take
$\lbs=\all$ and $\lepochs=1$ which corresponds exactly to
\fedsgdshort. For a client with $n_k$ local examples, the number of
local updates per round is given by $u_k = \lepochs\frac{n_k}{\lbs}$;
Complete pseudo-code is given in Algorithm~\ref{alg:fedavg}.

\setlength{\pw}{1.6in}
\begin{figure}[t]
\centering
\includegraphics[width=\pw]{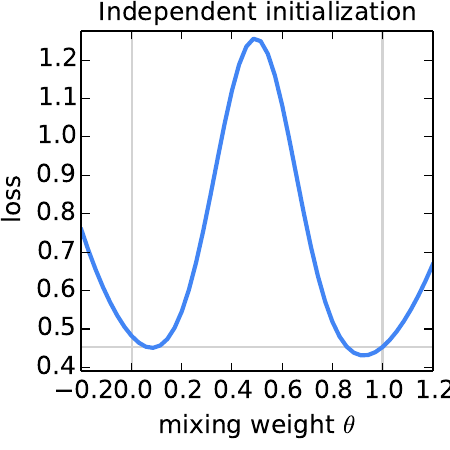}
\includegraphics[width=\pw]{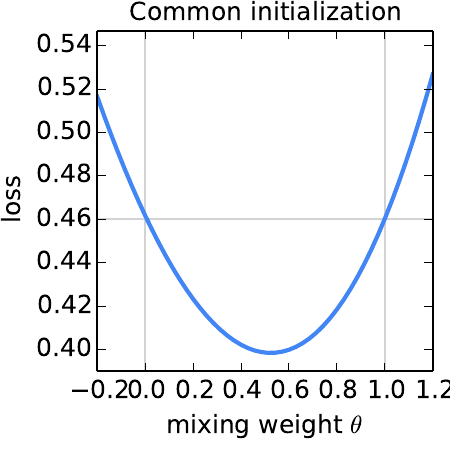} 
\mycaptionof{figure}{%
  The loss on the full MNIST training set for models generated by
  averaging the parameters of two models $w$ and $w'$ using $\theta w
  + (1-\theta) w'$ for 50 evenly spaced values $\theta \in [-0.2,
  1.2]$.The models $w$ and $w'$ were trained using SGD on
  different small datasets.
For the left plot, $w$ and $w'$ were initialized using different
random seeds; for the right plot, a shared seed was used. Note the
different $y$-axis scales. The horizontal line gives the best loss
achieved by $w$ or $w'$ (which were quite close, corresponding
  to the vertical lines at $\theta=0$ and $\theta=1$). With shared
initialization, averaging the models produces a significant reduction
in the loss on the total training set (much better than the loss of
either parent model).  }
\label{fig:mountain}
\end{figure}

For general non-convex objectives, averaging models in
parameter space could produce an arbitrarily bad model.  Following the
approach of \citet{goodfellow15qualitatively}, we see exactly this bad
behavior when we average two MNIST digit-recognition
models\footnote{We use the ``\tnn'' multi-layer perceptron described
  in Section~\ref{sec:models}.}  trained from different initial
conditions (Figure~\ref{fig:mountain}, left). For this figure, the
parent models $w$ and $w'$ were each trained on non-overlapping IID
samples of 600 examples from the MNIST training set. Training was via
SGD with a fixed learning rate of 0.1 for 240 updates on minibatches
of size 50 (or $\lepochs=20$ passes over the mini-datasets of
  size 600).  This is approximately the amount of training where the
models begin to overfit their local datasets.

\begin{algorithm}[t]
\begin{algorithmic}
\SUB{Server executes:}
   \STATE initialize $w_0$
   \FOR{each round $t = 1, 2, \dots$}
     \STATE $m \leftarrow \max(\clientfrac\cdot K, 1)$
     \STATE $S_t \leftarrow$ (random set of $m$ clients)
     \FOR{each client $k \in S_t$ \textbf{in parallel}}
       \STATE $w_{t+1}^k \leftarrow \text{ClientUpdate}(k, w_t)$ 
     \ENDFOR
     \STATE $m_t \leftarrow \sum_{k \in S_t} n_k$
     \STATE $w_{t+1} \leftarrow \sum_{k \in S_t} \frac{n_k}{m_t} w_{t+1}^k$\ \ %
         // \emph{Erratum}\footnotemark
   \ENDFOR
   \STATE

 \SUB{ClientUpdate($k, w$):}\ \ \  // \emph{Run on client $k$}
  \STATE $\mathcal{B} \leftarrow$ (split $\pp_k$ into batches of size $\lbs$)
  \FOR{each local epoch $i$ from $1$ to $\lepochs$}
    \FOR{batch $b \in \mathcal{B}$}
      \STATE $w \leftarrow w - \eta \grad \loss(w; b)$
    \ENDFOR
 \ENDFOR
 \STATE return $w$ to server
\end{algorithmic}
\mycaptionof{algorithm}{\fedavglong. The $\nc$
  clients are indexed by $k$; $\lbs$ is the local minibatch size,
  $\lepochs$ is the number of local epochs, and $\eta$ is the learning
  rate.}\label{alg:fedavg}
\end{algorithm}

Recent work indicates that in practice, the loss surfaces of
sufficiently over-parameterized NNs are surprisingly well-behaved and
in particular less prone to bad local minima than previously thought
\citep{dauphin14identifying,goodfellow15qualitatively,Choromanska15losssufaces}. And
indeed, when we start two models \emph{from the same random
  initialization} and then again train each independently on a
different subset of the data (as described above), we find that naive
parameter averaging works surprisingly well
(Figure~\ref{fig:mountain}, right): the average of these two models,
$\h w + \h w'$, achieves significantly lower loss on the full MNIST
training set than the best model achieved by training on either of the
small datasets independently. While Figure~\ref{fig:mountain} starts
from a random initialization, note a shared starting model $w_t$ is
used for each round of \fedavgshort, and so the same intuition applies.

\noaistats{The success of dropout training also provides some intuition for the
success of our model averaging scheme; dropout training can be
interpreted as averaging models of \emph{different} architectures
which share parameters, and the inference-time scaling of the model
parameters is analogous to the model averaging used in \fedavgshort
\citep{srivastava14dropout}.}

\section{Experimental Results} 
\label{sec:models} 
We are motivated by both image classification and language modeling
tasks where good models can greatly enhance the usability of mobile
devices. For each of these tasks we first picked a proxy dataset of
modest enough size that we could thoroughly investigate the
hyperparameters of the \fedavgshort algorithm.  While each individual
training run is relatively small, we trained over 2000 individual
models for these experiments. We then present results on the benchmark
CIFAR-10 image classification task. Finally, to demonstrate the
effectiveness of \fedavg on a real-world problem with a natural
partitioning of the data over clients, we evaluate on a large language
modeling task.

\footnotetext{Earlier versions of this paper incorrectly indicated summation over all $K$ clients here.}
Our initial study includes three model families on two datasets. The
first two are for the MNIST digit recognition task
\citep{lecun98gradientbased}:
1) A simple multilayer-perceptron with 2-hidden layers with 200 units
each using ReLu activations (199,210 total parameters), which we refer
to as the MNIST \tnn.
2) A CNN with two 5x5 convolution layers (the first with 32 channels,
the second with 64, each followed with 2x2 max pooling), a fully
connected layer with 512 units and ReLu activation, and a final
softmax output layer (1,663,370 total parameters).
To study federated optimization, we also need to specify how the data
is distributed over the clients. We study two ways of partitioning the
MNIST data over clients: \textbf{IID}, where the data is shuffled, and then
partitioned into 100 clients each receiving 600 examples, and
\textbf{Non-IID}, where we first \emph{sort} the data by digit label,
divide it into 200 shards of size 300, and assign each of 100 clients
2 shards. This is a pathological non-IID partition of
the data, as most clients will only have examples of two
digits, letting us explore the degree to which our algorithms
will break on highly non-IID data. Both of these partitions are balanced, however.\footnote{We performed additional experiments on unbalanced versions of these datasets, and found them to in fact be slightly easier for \fedavg.}

For language modeling, we built a dataset from \emph{The Complete
  Works of William Shakespeare} \citep{shakespeare}. We construct a
client dataset for each speaking role in each play with at least two
lines. This produced a dataset with 1146 clients. For each client, we
split the data into a set of training lines (the first 80\% of lines
for the role), and test lines (the last 20\%, rounded up to at least
one line). The resulting dataset has 3,564,579 characters in the
training set, and 870,014 characters\footnote{We always use character
  to refer to a one byte string, and use role to refer to a part in
  the play.}  in the test set. This data is substantially unbalanced,
with many roles having only a few lines, and a few with a large number
of lines.
Further, observe the test set is not a random sample of lines, but is
temporally separated by the chronology of each play. Using an
identical train/test split, we also form a balanced and IID version of
the dataset, also with 1146 clients.

On this data we train a stacked character-level LSTM language model,
which after reading each character in a line, predicts the next
character \citep{kim15character}.  The model takes a series of
characters as input and embeds each of these into a learned 8
dimensional space. The embedded characters are then processed through
2 LSTM layers, each with 256 nodes. Finally the output of the second
LSTM layer is sent to a softmax output layer with one node per
character. The full model has 866,578 parameters, and we trained using
an unroll length of 80 characters.

\newcommand*{\threeemdash}{\rule[0.5ex]{3em}{0.55pt}}
\begin{table}[t]
  \mycaptionof{table}{%
    Effect of the client fraction $\clientfrac$ on the MNIST \tnn with
    $\lepochs=1$ and CNN with $\lepochs=5$. Note $\clientfrac=0.0$
    corresponds to one client per round; since we use 100
      clients for the MNIST data, the rows correspond to 1, 10 20, 50,
      and 100 clients. Each table entry gives the number of rounds of
    communication necessary to achieve a test-set accuracy of
    \targetTNN for the \tnn and \targetCNN for the CNN, along with the
    speedup relative to the $\clientfrac=0$ baseline. Five runs with
    the large batch size did not reach the target accuracy in the
    allowed time.}\label{table:cfcnn}

\begin{center}
{\fontsize{7.5pt}{7.5pt}\selectfont
\begin{sc}
\begin{tabular}{l@{\ \ \ }r@{ }rr@{ }rr@{ }rr@{ }r}
 \hline
  \multicolumn{5}{l}{\T \textbf{\tnn} $\quad$ \ \ \ \threeemdash\ IID \threeemdash}
   & \multicolumn{4}{c}{\threeemdash Non-IID \threeemdash}     \\
 \multicolumn{1}{c}{$\clientfrac$\T}   %
   &  \multicolumn{2}{c}{$\lbs=\all$} 
   &  \multicolumn{2}{c}{$\lbs=10$} 
   &  \multicolumn{2}{c}{$\lbs=\all$} 
   &  \multicolumn{2}{c}{$\lbs=10$} \\
\hline
0.0 &   1455 &             &   316 &            &  4278 &            &  3275 &     \T   \\  
0.1 &   1474 & $(1.0\xx)$  &    87 & $(3.6\xx)$ &  1796 & $(2.4\xx)$ &   664 & $(4.9\xx)$ \\  
0.2 &   1658 & $(0.9\xx)$  &    77 & $(4.1\xx)$ &  1528 & $(2.8\xx)$ &   619 & $(5.3\xx)$ \\  
0.5 &   ---  & (---)       &    75 & $(4.2\xx)$ &  ---  & (---)      &   443 & $(7.4\xx)$ \\  
1.0 &   ---  & (---)       &    70 & $(4.5\xx)$ &  ---  & (---)      &   380 & $(8.6\xx)$ \\  
\hline
\multicolumn{4}{l}{\textbf{CNN}, $\lepochs=5$ \T} &\\
\hline
0.0 &  387&          & 50 &  & 1181 &  & 956 &  \T \\
0.1 &  339& $(1.1\xx)$ & 18 & $(2.8\xx)$ & 1100 & $(1.1\xx)$ & 206 & $(4.6\xx)$ \\
0.2 &  337& $(1.1\xx)$ & 18 & $(2.8\xx)$ &  978 & $(1.2\xx)$ & 200 & $(4.8\xx)$ \\
0.5 &  164& $(2.4\xx)$ & 18 & $(2.8\xx)$ & 1067 & $(1.1\xx)$ & 261 & $(3.7\xx)$ \\
1.0 &  246& $(1.6\xx)$ & 16 & $(3.1\xx)$ &  --- & (---) &  97 & $(9.9\xx)$ \\
\hline
\end{tabular}
\end{sc}
}
\end{center}
\end{table}

\begin{table}[t]
\begin{center}
  \mycaptionof{table}{Number of communication rounds to reach a target
    accuracy for \fedavgshort, versus \fedsgdshort (first row,
    $\lepochs=1$ and $\lbs=\infty$). The $u$ column gives
    $u=\lepochs n / (\nc \lbs)$, the expected number of updates per
    round.  }
\label{table:speedupshort}

{\fontsize{7.5pt}{7.5pt}\selectfont
\begin{sc}
\begin{tabular}{rrrrr@{ }rr@{ }r}
\hline
& \multicolumn{7}{l}{\hspace{-0.1in}\textbf{MNIST CNN}, \targetCNN accuracy \T}\\
\T \textbf{CNN} & $\lepochs$ & $\lbs$ & $u$
 & \multicolumn{2}{c}{IID} 
 & \multicolumn{2}{c}{Non-IID} \\
\hline
\fedsgdshort &    1 &$\all$&    1 &    626 &             &    483 &          \T \\
\fedavgshort &    5 &$\all$&    5 &    179 & $(   3.5\xx)$ &   1000 & $(   0.5\xx)$ \\
\fedavgshort &    1 &   50 &   12 &     65 & $(   9.6\xx)$ &    600 & $(   0.8\xx)$ \\
\fedavgshort &   20 &$\all$&   20 &    234 & $(   2.7\xx)$ &    672 & $(   0.7\xx)$ \\
\fedavgshort &    1 &   10 &   60 &     34 & $(  18.4\xx)$ &    350 & $(   1.4\xx)$ \\
\fedavgshort &    5 &   50 &   60 &     29 & $(  21.6\xx)$ &    334 & $(   1.4\xx)$ \\
\fedavgshort &   20 &   50 &  240 &     32 & $(  19.6\xx)$ &    426 & $(   1.1\xx)$ \\
\fedavgshort &    5 &   10 &  300 &     20 & $(  31.3\xx)$ &    229 & $(   2.1\xx)$ \\
\fedavgshort &   20 &   10 & 1200 &     18 & $(  34.8\xx)$ &    173 & $(   2.8\xx)$ \\
\hline
& \multicolumn{7}{l}{\hspace{-0.1in}\textbf{Shakespeare LSTM}, \targetLSTM accuracy \T}\\
\T \textbf{LSTM} & $\lepochs$ & $\lbs$ & $u$
 & \multicolumn{2}{c}{IID} 
 & \multicolumn{2}{c}{Non-IID} \\
\hline
\fedsgdshort &    1 & $\infty$ &  1.0 &   2488 &               &   3906 &           \T  \\
\fedavgshort &    1 &       50 &  1.5 &   1635 & $(   1.5\xx)$ &    549 & $(   7.1\xx)$ \\
\fedavgshort &    5 & $\infty$ &  5.0 &    613 & $(   4.1\xx)$ &    597 & $(   6.5\xx)$ \\
\fedavgshort &    1 &       10 &  7.4 &    460 & $(   5.4\xx)$ &    164 & $(  23.8\xx)$ \\
\fedavgshort &    5 &       50 &  7.4 &    401 & $(   6.2\xx)$ &    152 & $(  25.7\xx)$ \\
\fedavgshort &    5 &       10 & 37.1 &    192 & $(  13.0\xx)$ &     41 & $(  95.3\xx)$ \\  
\hline
\end{tabular}
\end{sc}
}
\end{center}
\end{table}

SGD is sensitive to the tuning of the learning-rate parameter
$\eta$. The results reported here are based on training
over a sufficiently wide grid of learning rates (typically 11-13
values for $\eta$ on a multiplicative grid of resolution
$10^{\frac{1}{3}}$ or $10^{\frac{1}{6}}$). We checked to ensure the
best learning rates were in the middle of our grids, and that there
was not a significant difference between the best learning rates.
Unless otherwise noted, we plot metrics for the best performing rate
selected individually for each $x$-axis value.  
We find that the optimal learning rates do not vary too
much as a function of the other parameters. 

\paragraph{Increasing parallelism} 
We first experiment with the client fraction $\clientfrac$, which controls
the amount of multi-client parallelism. Table~\ref{table:cfcnn} shows
the impact of varying $\clientfrac$ for both MNIST models.
We report the number of communication rounds necessary to achieve a
target test-set accuracy.  To compute this, we construct a learning
curve for each combination of parameter settings, optimizing $\eta$ as
described above and then making each curve monotonically improving by
taking the best value of test-set accuracy achieved over all prior
rounds. We then calculate the number of rounds where the curve crosses
the target accuracy, using linear interpolation between the discrete
points forming the curve.  This is perhaps best understood by
reference to Figure~\ref{fig:testaccuracy}, where the gray lines show
the targets.

With $\lbs=\all$ (for MNIST processing all 600 client examples as a
single batch per round), there is only a small advantage in increasing
the client fraction. Using the smaller batch size $\lbs=10$ shows a
significant improvement in using $\clientfrac \ge 0.1$, especially in
the non-IID case. Based on these results, for most of the remainder of
our experiments we fix $\clientfrac = 0.1$, which strikes a good
balance between computational efficiency and convergence
rate. Comparing the number of rounds for the $\lbs=\all$ and $\lbs=10$
columns in Table~\ref{table:cfcnn} shows a dramatic speedup, which we
investigate next.

\paragraph{Increasing computation per client}
In this section, we fix $\clientfrac=0.1$, and add more computation
per client on each round, either decreasing $\lbs$, increasing
$\lepochs$, or both. Figure~\ref{fig:testaccuracy} demonstrates that
adding more local SGD updates per round can produce a dramatic
decrease in communication costs, and Table~\ref{table:speedupshort}
quantifies these speedups.
The expected number of updates per client per
round is $u = (\E[n_k] / \lbs) \lepochs = n \lepochs / (\nc \lbs)$,
where the expectation is over the draw of a random client $k$. We
order the rows in each section of Table~\ref{table:speedupshort} by
this statistic. We see that increasing $u$ by varying both $\lepochs$
and $\lbs$ is effective. As long as $\lbs$ is large enough to take
full advantage of available parallelism on the client hardware, there
is essentially no cost in computation time for lowering it, and so in
practice this should be the first parameter tuned.

\setlength{\pw}{1.6in}
\begin{figure}[t]
\begin{center}
 \includegraphics[width=\pw]{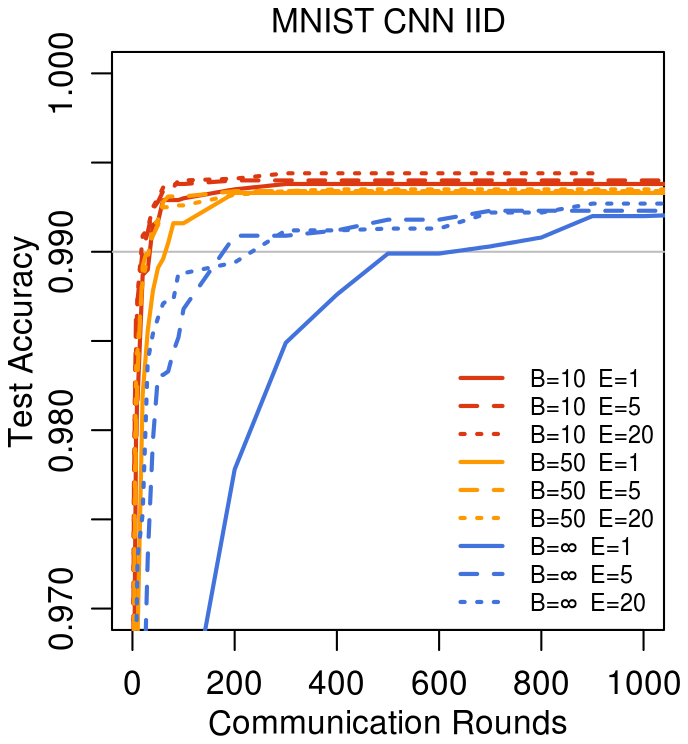}
 \includegraphics[width=\pw]{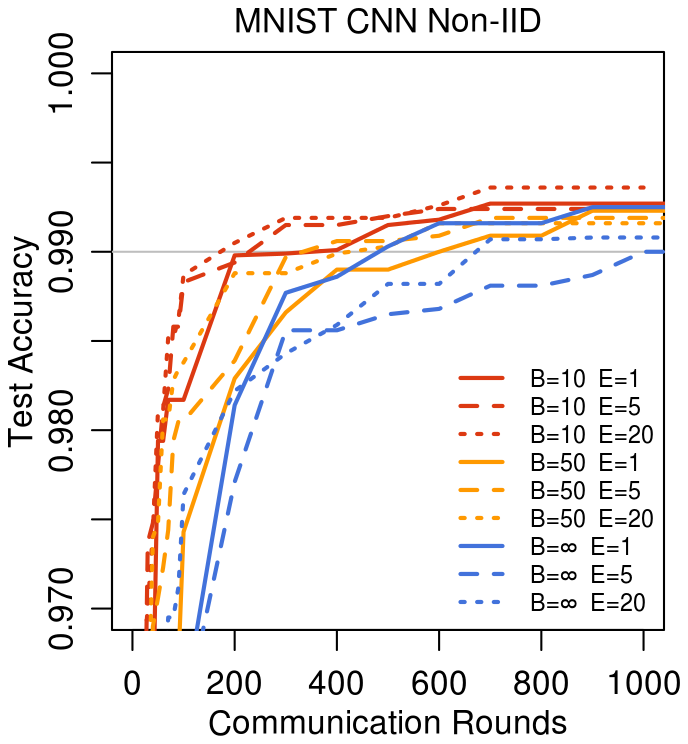}

 \vspace{0.1in}
\includegraphics[width=\pw]{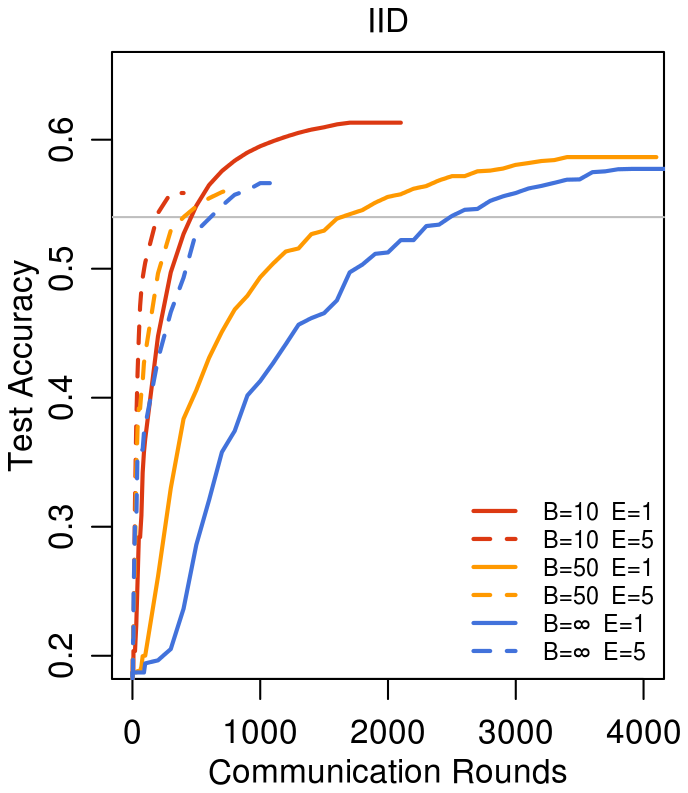}
\includegraphics[width=\pw]{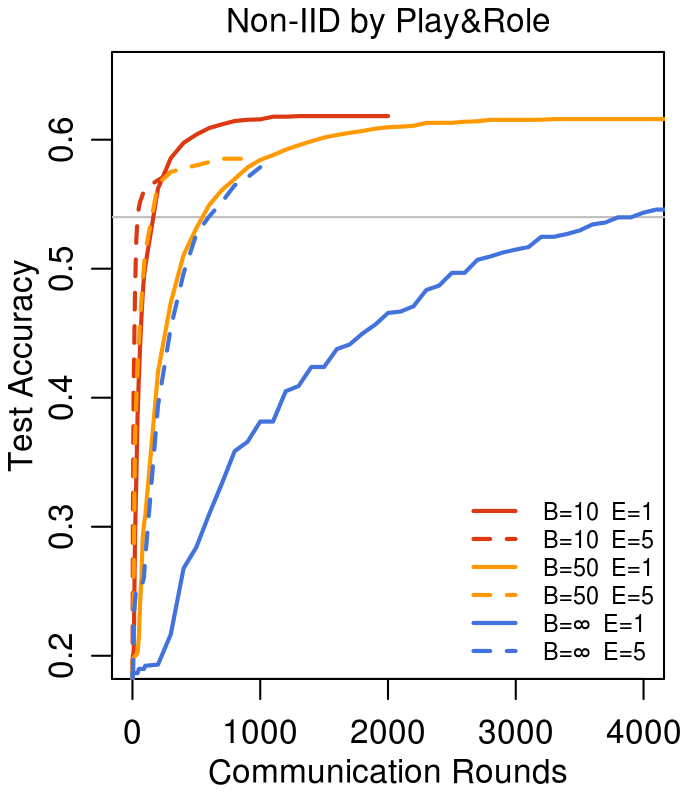}
\mycaptionof{figure}{Test set accuracy vs. communication rounds for
  the MNIST CNN (IID and then pathological non-IID) and
  Shakespeare LSTM (IID and then by Play\&Role) with $\clientfrac=0.1$
  and optimized $\eta$. The gray lines show the target accuracies used
  in Table~\ref{table:speedupshort}. Plots for the \tnn are given as
  Figure~\ref{fig:mnisttnn} in Appendix~\ref{app:fig}.}
\label{fig:testaccuracy}
\end{center}
\end{figure}

For the IID partition of the MNIST data, using more computation per
client decreases the number of rounds to reach the target accuracy by
$35\times$ for the CNN and $46\times$ for the \tnn (see
Table~\ref{table:tnn} in Appendix~\ref{app:fig} for details for the
\tnn). The speedups for the pathologically partitioned non-IID data are
smaller, but still substantial ($2.8$ -- $3.7\times$). It is
impressive that averaging provides \emph{any} advantage (vs. actually
diverging) when we naively average the parameters of models trained on
entirely different pairs of digits. Thus, we view this as strong
evidence for the robustness of this approach.

The unbalanced and non-IID distribution of the Shakespeare (by
role in the play) is much more representative of the kind of data
distribution we expect for real-world applications. Encouragingly, for
this problem learning on the non-IID and unbalanced data is actually
much easier (a $95\times$ speedup vs $13\times$ for the balanced IID
data); we conjecture this is largely due to the fact some roles have
relatively large local datasets, which makes increased local training
particularly valuable.

For all three model classes, \fedavg converges to a
higher level of test-set accuracy than the baseline \fedsgd models.
This trend continues even if the lines are extended beyond the plotted
ranges. For example, for the CNN the $\lbs=\all, \lepochs=1$
\fedsgdshort model eventually reaches 99.22\% accuracy after 1200
rounds (and had not improved further after 6000 rounds), while the
$\lbs=10, \lepochs=20$ \fedavgshort model reaches an accuracy of
99.44\% after 300 rounds. We conjecture that in addition to lowering
communication costs, model averaging produces a regularization benefit
similar to that achieved by dropout \citep{srivastava14dropout}.

We are primarily concerned with generalization performance, but
\fedavgshort is effective at optimizing the training loss as well,
even beyond the point where test-set accuracy plateaus. We observed
similar behavior for all three model classes, and present plots for
the MNIST CNN in Figure~\ref{fig:mnist-train-loss} in
Appendix~\ref{app:fig}.

\setlength{\pw}{1.6in}
\begin{figure}
\begin{center}
\includegraphics[width=\pw]{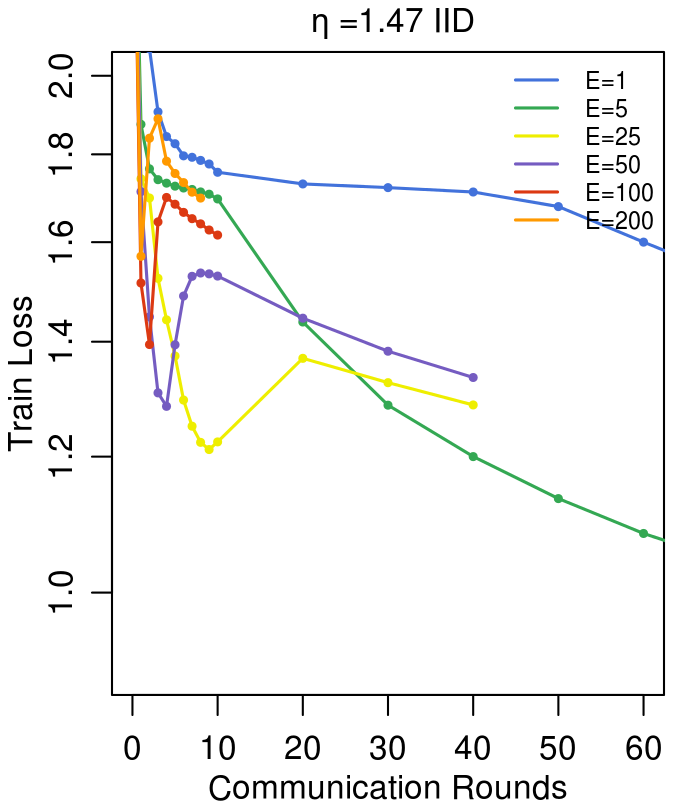}  
\includegraphics[width=\pw]{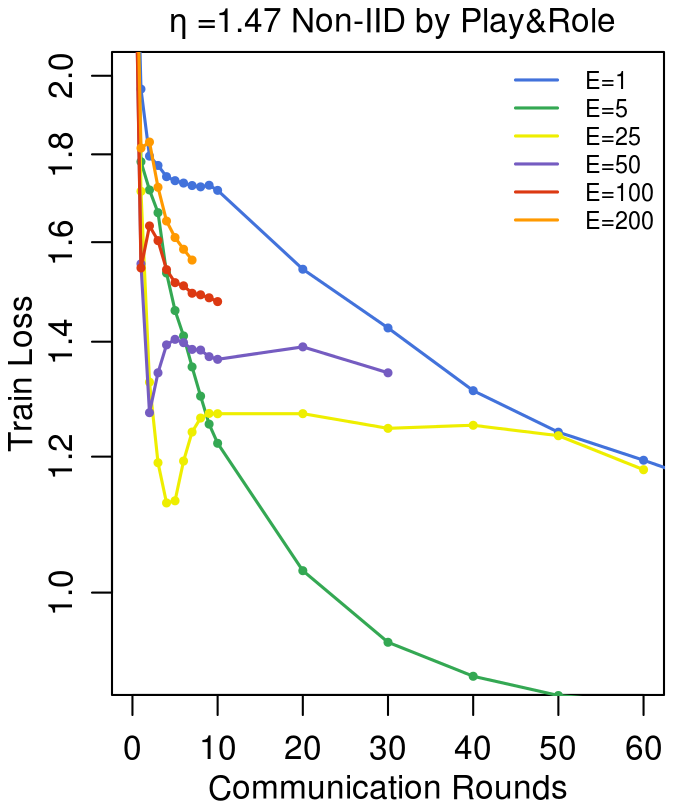} 
\vskip -0.05in
\mycaptionof{figure}{The effect of training for many local epochs (large
  $\lepochs$) between averaging steps, fixing $\lbs=10$ and
  $\clientfrac=0.1$ for the Shakespeare
  LSTM with a fixed learning rate $\eta=1.47$.}\label{fig:lstm-many-epochs}
\end{center}
\end{figure}

\paragraph{Can we over-optimize on the client datasets?}
The current model parameters only influence the optimization
performed in each \texttt{ClientUpdate} via initialization.  Thus, as
$\lepochs \rightarrow \infty$, at least for a convex problem
eventually the initial conditions should be irrelevant, and the global
minimum would be reached regardless of initialization. Even for a
non-convex problem, one might conjecture the algorithm would converge
to the same local minimum as long as the initialization was in the same
basin. That is, we would expect that while one round of averaging
might produce a reasonable model, additional rounds of communication
(and averaging) would not produce further improvements.

Figure~\ref{fig:lstm-many-epochs} shows the impact of large
$\lepochs$ during initial training on the Shakespeare LSTM
problem. Indeed, for very large numbers of local epochs, \fedavgshort
can plateau or diverge.\footnote{ Note that due to this behavior and
  because for large $\lepochs$ not all experiments for all learning
  rates were run for the full number of rounds, we report results for
  a fixed learning rate (which perhaps surprisingly was near-optimal
  across the range of $\lepochs$ parameters) and without forcing the
  lines to be monotonic.}
This result suggests that for some models, especially in the later
stages of convergence, it may be useful to decay the amount of local
computation per round (moving to smaller $\lepochs$ or larger $\lbs$)
in the same way decaying learning rates can be useful.
Figure~\ref{fig:cnn-many-epochs} in Appendix~\ref{app:fig} gives the
analogous experiment for the MNIST CNN. Interestingly, for this model
we see no significant degradation in the convergence rate for large
values of $\lepochs$. However, we see slightly better performance for
$\lepochs=1$ versus $\lepochs=5$ for the large-scale language modeling
task described below (see Figure~\ref{fig:accuracyvar} in
Appendix~\ref{app:fig}).

\begin{figure}[t]
  \begin{center}
    \includegraphics[width=3.2in]{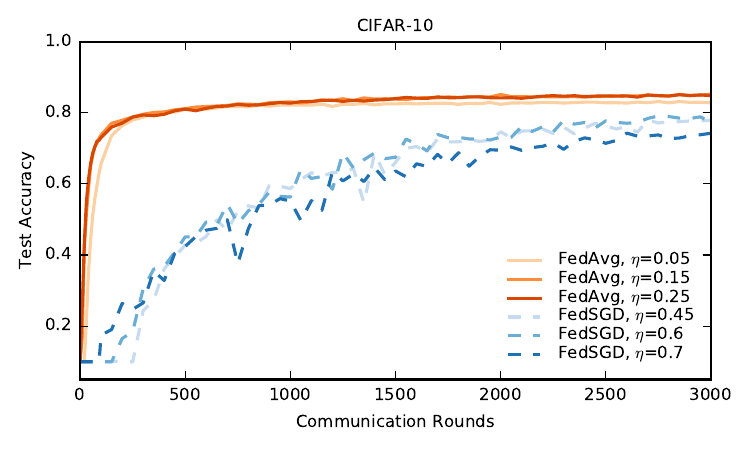} \\
    \vskip -0.05in \mycaptionof{figure}{Test accuracy versus
      communication for the CIFAR10 experiments. \fedsgd uses a
      learning-rate decay of 0.9934 per round; \fedavgshort uses
      $\lbs=50$, learning-rate decay of 0.99 per round, and
      $\lepochs=5$.}\label{fig:cifar}
 \end{center}
\end{figure}

\paragraph{CIFAR experiments}
We also ran experiments on the CIFAR-10 dataset
\citep{krizhevsky09cifar} to further validate \fedavg. The dataset
consists of 10 classes of 32x32 images with three RGB
channels. There are 50,000 training examples and 10,000 testing
examples, which we partitioned into 100 clients each containing 500
training and 100 testing examples; since there isn't a natural user
partitioning of this data, we considered the balanced and IID setting.
The model architecture was taken from the TensorFlow
tutorial \citep{tensorflowcifar}, which consists of two convolutional
layers followed by two fully connected layers and then a linear
transformation layer to produce logits, for a total of about $10^6$ parameters.
Note that state-of-the-art approaches have achieved a test accuracy of
96.5\% \citep{graham14fractional} for CIFAR; nevertheless, the
standard model we use is sufficient for our needs, as our goal is to
evaluate our optimization method, not achieve the best possible
accuracy on this task.
The images are preprocessed as part of the training input pipeline,
which consists of cropping the images to 24x24, randomly flipping
left-right and adjusting the contrast, brightness and whitening.

For these experiments, we considered an additional baseline, standard
SGD training on the full training set (no user partitioning), using
minibatches of size 100. We achieved an ~86\% test accuracy after
197,500 minibatch updates (each minibatch update requires a
communication round in the federated setting).
\fedavg achieves a similar test accuracy of 85\% after only 2,000
communication rounds. For all algorithms, we tuned a learning-rate
decay parameter in addition to the initial learning rate.
Table~\ref{table:cifar} gives the number of communication rounds for
baseline SGD, \fedsgd, and \fedavg to reach three different accuracy
targets, and Figure~\ref{fig:cifar} gives learning-rate curves for
\fedavgshort versus \fedsgdshort.

By running experiments with minibatches of size $\lbs=50$ for both SGD
and \fedavg, we can also look at accuracy as a function of the number
of such minibatch gradient calculations. We expect SGD to do
better here, because a sequential step is taken after each minibatch
computation. However, as Figure~\ref{fig:cifarbatches} in the appendix
shows, for modest values of $\clientfrac$ and $\lepochs$, \fedavg makes
a similar amount of progress per minibatch computation. Further, we
see that both standard SGD and \fedavgshort with only one client per
round $(\clientfrac = 0)$, demonstrate significant oscillations in
accuracy, whereas averaging over more clients smooths this out.

\newcommand{\acc}[1]{\multicolumn{2}{c}{#1\%}}
\begin{table}[t]
  \mycaptionof{table}{Number of rounds and speedup relative to baseline SGD to reach a target test-set accuracy on CIFAR10. SGD used a minibatch size of 100. \fedsgd and \fedavg used $\clientfrac=0.1$, with \fedavg using $\lepochs=5$ and $\lbs=50$.
  }\label{table:cifar}
  {\fontsize{7.5pt}{7.5pt}\selectfont   
    \begin{sc}
  \begin{tabular}{lHHr@{ }rr@{ }rr@{ }r}
    \hline
    Acc.      &  &       & \acc{80}         & \acc{82}           &  \acc{85}   \T \\
    \hline
    \algfont{SGD} \T & 2000 & (---)     & 18000 & (---)     & 31000 & (---)     & 99000 & (---) \\
    \fedsgd          & 910  & (2.2\xx)  &  3750 & (4.8\xx)  &  6600 & (4.7\xx)  &   n/a & (---) \\
    \fedavg          & 40   & (50.0\xx) &   280 & (64.3\xx) &   630 & (49.2\xx) & 2000  & (49.5\xx) \\
    \hline
  \end{tabular}
  \end{sc}
  }
\end{table}

\paragraph{Large-scale LSTM experiments}

We ran experiments on a large-scale next-word prediction task to
demonstrate the effectiveness of our approach on a real-world
problem. Our training dataset consists $10$ million public posts from
a large social network. We grouped the posts by author, for a total of
over 500,000 clients. This dataset is a realistic proxy for the type
of text entry data that would be present on a user's mobile device. We
limited each client dataset to at most 5000 words, and report accuracy
(the fraction of the data where the highest predicted probability was
on the correct next word, out of 10000 possibilities) on a test set of
$1e5$ posts from different (non-training) authors.
Our model is a 256 node LSTM on a vocabulary of 10,000
words. The input and output embeddings for each word were of dimension
192, and co-trained with the model; there are 4,950,544 parameters in
all. We used an unroll of 10 words.

\begin{figure}[t]
  \begin{center}
    \includegraphics[width=3.2in]{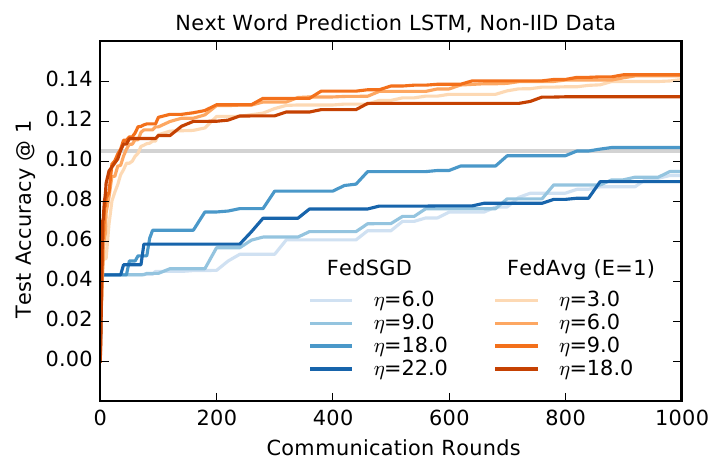} \\
    \vskip -0.05in
    \mycaptionof{figure}{Monotonic learning curves for the large-scale language
      model word LSTM.}\label{fig:wordlstm}
 \end{center}
\end{figure}

These experiments required significant computational resources 
and so we did not explore hyper-parameters as
thoroughly: all runs trained on 200 clients per round; \fedavgshort
used $\lbs=8$ and $\lepochs=1$. We explored a variety of learning
rates for \fedavgshort and the baseline \fedsgd. 
Figure~\ref{fig:wordlstm} shows monotonic learning curves for the best
learning rates. \fedsgdshort with $\eta = 18.0$ required 820 rounds to
reach 10.5\% accuracy, while \fedavgshort with $\eta=9.0$ reached an
accuracy of 10.5\% in only 35 communication rounds ($23\times$ fewer
then \fedsgdshort).
We observed lower variance in test accuracy for \fedavgshort, see
Figure~\ref{fig:accuracyvar} in Appendix~\ref{app:fig}. This figure also
include results for $\lepochs=5$, which performed
slightly worse than $\lepochs=1$.

\section{Conclusions and Future Work}
Our experiments show that federated learning can be made practical, as
\fedavg trains high-quality models using relatively few rounds of
communication, as demonstrated by results on a variety of model
architectures: a multi-layer perceptron, two different convolutional
NNs, a two-layer character LSTM, and a large-scale word-level
LSTM.

While federated learning offers many practical privacy benefits,
providing stronger guarantees via differential
privacy~\citep{dwork14book,duchi14privacy,abadi16dpdl}, secure
multi-party computation \citep{goryczka13secure}, or their combination
is an interesting direction for future work. Note that both classes of
techniques apply most naturally to synchronous algorithms like
\fedavg.\footnote{Subsequent to this work, \citet{bonawitz16secaggworkshop} introduced an efficient secure aggregation protocol for federated learning, and \citet{konecny15commworkshop} presented algorithms for further decreasing communication costs.}

\noaistats{In order to keep
the scope of algorithms explored tractable, we limited ourselves to
building on vanilla SGD. Investigating the compatibility of our
approach with other optimization algorithms such as momentum~\citep{sutskever13importance}, 
AdaGrad~\citep{mcmahan10boundopt,duchi11adaptivejournal} and
ADAM~\citep{kingma15adam}, as well as with changes in model structure
that can aid optimization, such as dropout~\citep{srivastava14dropout}
and batch-normalization~\citep{ioffe16batchnorm}, are another natural
direction for future work.}

\clearpage
\bibliography{../new,../my_pubs}

\begin{thebibliography}{46}
\providecommand{\natexlab}[1]{#1}
\providecommand{\url}[1]{\texttt{#1}}
\expandafter\ifx\csname urlstyle\endcsname\relax
  \providecommand{\doi}[1]{doi: #1}\else
  \providecommand{\doi}{doi: \begingroup \urlstyle{rm}\Url}\fi

\bibitem[Abadi et~al.(2016)Abadi, Chu, Goodfellow, McMahan, Mironov, Talwar,
  and Zhang]{abadi16dpdl}
Martin Abadi, Andy Chu, Ian Goodfellow, Brendan McMahan, Ilya Mironov, Kunal
  Talwar, and Li~Zhang.
\newblock Deep learning with differential privacy.
\newblock In \emph{23rd ACM Conference on Computer and Communications Security
  (ACM CCS)}, 2016.

\bibitem[Anderson(2015)]{pew15deviceownership}
Monica Anderson.
\newblock Technology device ownership: 2015.
\newblock
  \url{http://www.pewinternet.org/2015/10/29/technology-device-ownership-2015/},
  2015.

\bibitem[Arjevani and Shamir(2015)]{arjevani15commcomplexity}
Yossi Arjevani and Ohad Shamir.
\newblock Communication complexity of distributed convex learning and
  optimization.
\newblock In \emph{Advances in Neural Information Processing Systems 28}. 2015.

\bibitem[Balcan et~al.(2012)Balcan, Blum, Fine, and
  Mansour]{balcan12distributed}
Maria-Florina Balcan, Avrim Blum, Shai Fine, and Yishay Mansour.
\newblock Distributed learning, communication complexity and privacy.
\newblock \emph{arXiv preprint arXiv:1204.3514}, 2012.

\bibitem[Bengio et~al.(2003)Bengio, Ducharme, Vincent, and
  Janvin]{bengio03neural}
Yoshua Bengio, R{\'e}jean Ducharme, Pascal Vincent, and Christian Janvin.
\newblock A neural probabilistic language model.
\newblock \emph{J. Mach. Learn. Res.}, 2003.

\bibitem[Bonawitz et~al.(2016)Bonawitz, Ivanov, Kreuter, Marcedone, McMahan,
  Patel, Ramage, Segal, and Seth]{bonawitz16secaggworkshop}
Keith Bonawitz, Vladimir Ivanov, Ben Kreuter, Antonio Marcedone, H.~Brendan
  McMahan, Sarvar Patel, Daniel Ramage, Aaron Segal, and Karn Seth.
\newblock Practical secure aggregation for federated learning on user-held
  data.
\newblock In \emph{NIPS Workshop on Private Multi-Party Machine Learning},
  2016.

\bibitem[Chaum(1981)]{chaum81untraceable}
David~L. Chaum.
\newblock Untraceable electronic mail, return addresses, and digital
  pseudonyms.
\newblock \emph{Commun. ACM}, 24\penalty0 (2), 1981.

\bibitem[Chen et~al.(2016)Chen, Monga, Bengio, and
  Jozefowicz]{chen16revisiting}
Jianmin Chen, Rajat Monga, Samy Bengio, and Rafal Jozefowicz.
\newblock Revisiting distributed synchronous sgd.
\newblock In \emph{ICLR Workshop Track}, 2016.

\bibitem[Choromanska et~al.(2015)Choromanska, Henaff, Mathieu, Arous, and
  LeCun]{Choromanska15losssufaces}
Anna Choromanska, Mikael Henaff, Micha{\"{e}}l Mathieu, G{\'{e}}rard~Ben Arous,
  and Yann LeCun.
\newblock The loss surfaces of multilayer networks.
\newblock In \emph{{AISTATS}}, 2015.

\bibitem[Corrado(2015)]{corrado15computer}
Greg Corrado.
\newblock Computer, respond to this email.
\newblock
  \url{http://googleresearch.blogspot.com/2015/11/computer-respond-to-this-email.html},
  November 2015.

\bibitem[Dauphin et~al.(2014)Dauphin, Pascanu, G{\"{u}}l{\c{c}}ehre, Cho,
  Ganguli, and Bengio]{dauphin14identifying}
Yann~N. Dauphin, Razvan Pascanu, {\c{C}}aglar G{\"{u}}l{\c{c}}ehre, KyungHyun
  Cho, Surya Ganguli, and Yoshua Bengio.
\newblock Identifying and attacking the saddle point problem in
  high-dimensional non-convex optimization.
\newblock In \emph{{NIPS}}, 2014.

\bibitem[Dean et~al.(2012)Dean, Corrado, Monga, Chen, Devin, Le, Mao, Ranzato,
  Senior, Tucker, Yang, and Ng]{dean12large}
Jeffrey Dean, Greg~S. Corrado, Rajat Monga, Kai Chen, Matthieu Devin, Quoc~V.
  Le, Mark~Z. Mao, Marc'Aurelio Ranzato, Andrew Senior, Paul Tucker, Ke~Yang,
  and Andrew~Y. Ng.
\newblock Large scale distributed deep networks.
\newblock In \emph{NIPS}, 2012.

\bibitem[Duchi et~al.(2014)Duchi, Jordan, and Wainwright]{duchi14privacy}
John Duchi, Michael~I. Jordan, and Martin~J. Wainwright.
\newblock Privacy aware learning.
\newblock \emph{Journal of the Association for Computing Machinery}, 2014.

\bibitem[Dwork and Roth(2014)]{dwork14book}
Cynthia Dwork and Aaron Roth.
\newblock \emph{The Algorithmic Foundations of Differential Privacy}.
\newblock Foundations and Trends in Theoretical Computer Science. Now
  Publishers, 2014.

\bibitem[Fercoq et~al.(2014)Fercoq, Qu, Richt{\'a}rik, and
  Tak{\'a}c]{fercoq14fast}
Olivier Fercoq, Zheng Qu, Peter Richt{\'a}rik, and Martin Tak{\'a}c.
\newblock Fast distributed coordinate descent for non-strongly convex losses.
\newblock In \emph{Machine Learning for Signal Processing (MLSP), 2014 IEEE
  International Workshop on}, 2014.

\bibitem[Goodfellow et~al.(2016)Goodfellow, Bengio, and
  Courville]{goodfellow16deeplearning}
Ian Goodfellow, Yoshua Bengio, and Aaron Courville.
\newblock Deep learning.
\newblock Book in preparation for MIT Press, 2016.

\bibitem[Goodfellow et~al.(2015)Goodfellow, Vinyals, and
  Saxe]{goodfellow15qualitatively}
Ian~J. Goodfellow, Oriol Vinyals, and Andrew~M. Saxe.
\newblock Qualitatively characterizing neural network optimization problems.
\newblock In \emph{ICLR}, 2015.

\bibitem[Goryczka et~al.(2013)Goryczka, Xiong, and Sunderam]{goryczka13secure}
Slawomir Goryczka, Li~Xiong, and Vaidy Sunderam.
\newblock Secure multiparty aggregation with differential privacy: A
  comparative study.
\newblock In \emph{Proceedings of the Joint EDBT/ICDT 2013 Workshops}, 2013.

\bibitem[Graham(2014)]{graham14fractional}
Benjamin Graham.
\newblock Fractional max-pooling.
\newblock \emph{CoRR}, abs/1412.6071, 2014.
\newblock URL \url{http://arxiv.org/abs/1412.6071}.

\bibitem[Hochreiter and Schmidhuber(1997)]{hochreiter97lstm}
Sepp Hochreiter and J\"{u}rgen Schmidhuber.
\newblock Long short-term memory.
\newblock \emph{Neural Computation}, 9\penalty0 (8), November 1997.

\bibitem[Ioffe and Szegedy(2015)]{ioffe16batchnorm}
Sergey Ioffe and Christian Szegedy.
\newblock Batch normalization: Accelerating deep network training by reducing
  internal covariate shift.
\newblock In \emph{ICML}, 2015.

\bibitem[Kim et~al.(2015)Kim, Jernite, Sontag, and Rush]{kim15character}
Yoon Kim, Yacine Jernite, David Sontag, and Alexander~M. Rush.
\newblock Character-aware neural language models.
\newblock \emph{CoRR}, abs/1508.06615, 2015.

\bibitem[Kone\v{c}n\'{y} et~al.(2016)Kone\v{c}n\'{y}, McMahan, Yu, Richtarik,
  Suresh, and Bacon]{konecny15commworkshop}
Jakub Kone\v{c}n\'{y}, H.~Brendan McMahan, Felix~X. Yu, Peter Richtarik,
  Ananda~Theertha Suresh, and Dave Bacon.
\newblock Federated learning: Strategies for improving communication
  efficiency.
\newblock In \emph{NIPS Workshop on Private Multi-Party Machine Learning},
  2016.

\bibitem[Krizhevsky(2009)]{krizhevsky09cifar}
Alex Krizhevsky.
\newblock Learning multiple layers of features from tiny images.
\newblock Technical report, 2009.

\bibitem[Krizhevsky et~al.(2012)Krizhevsky, Sutskever, and
  Hinton]{krizhevsky12imagenet}
Alex Krizhevsky, Ilya Sutskever, and Geoffrey~E. Hinton.
\newblock Imagenet classification with deep convolutional neural networks.
\newblock In \emph{NIPS}. 2012.

\bibitem[LeCun et~al.(1998)LeCun, Bottou, Bengio, and
  Haffner]{lecun98gradientbased}
Y.~LeCun, L.~Bottou, Y.~Bengio, and P.~Haffner.
\newblock Gradient-based learning applied to document recognition.
\newblock \emph{Proceedings of the IEEE}, 86\penalty0 (11), 1998.

\bibitem[Ma et~al.(2015)Ma, Smith, Jaggi, Jordan, Richt{\'a}rik, and
  Tak{\'a}{\v{c}}]{chenxin15cocoa}
Chenxin Ma, Virginia Smith, Martin Jaggi, Michael~I Jordan, Peter
  Richt{\'a}rik, and Martin Tak{\'a}{\v{c}}.
\newblock Adding vs. averaging in distributed primal-dual optimization.
\newblock In \emph{ICML}, 2015.

\bibitem[McDonald et~al.(2010)McDonald, Hall, and Mann]{mcdonald10distributed}
Ryan McDonald, Keith Hall, and Gideon Mann.
\newblock Distributed training strategies for the structured perceptron.
\newblock In \emph{NAACL HLT}, 2010.

\bibitem[Neverova et~al.(2016)Neverova, Wolf, Lacey, Fridman, Chandra,
  Barbello, and Taylor]{neverova16motion}
Natalia Neverova, Christian Wolf, Griffin Lacey, Lex Fridman, Deepak Chandra,
  Brandon Barbello, and Graham~W. Taylor.
\newblock Learning human identity from motion patterns.
\newblock \emph{{IEEE} Access}, 4:\penalty0 1810--1820, 2016.

\bibitem[Poushter(2016)]{pew16smartphone}
Jacob Poushter.
\newblock Smartphone ownership and internet usage continues to climb in
  emerging economies.
\newblock
  \href{http://www.pewglobal.org/2016/02/22/smartphone-ownership-and-internet-usage-continues-to-climb-in-emerging-economies/}{Pew
  Research Center Report}, 2016.

\bibitem[Povey et~al.(2015)Povey, Zhang, and Khudanpur]{povey15parallel}
Daniel Povey, Xiaohui Zhang, and Sanjeev Khudanpur.
\newblock Parallel training of deep neural networks with natural gradient and
  parameter averaging.
\newblock In \emph{ICLR Workshop Track}, 2015.

\bibitem[Shakespeare()]{shakespeare}
William Shakespeare.
\newblock {The Complete Works of William Shakespeare}.
\newblock Publically available at \url{https://www.gutenberg.org/ebooks/100}.

\bibitem[Shamir and Srebro(2014)]{shamir14distributed}
Ohad Shamir and Nathan Srebro.
\newblock Distributed stochastic optimization and learning.
\newblock In \emph{Communication, Control, and Computing (Allerton)}, 2014.

\bibitem[Shamir et~al.(2013)Shamir, Srebro, and Zhang]{shamir13dane}
Ohad Shamir, Nathan Srebro, and Tong Zhang.
\newblock Communication efficient distributed optimization using an approximate
  newton-type method.
\newblock \emph{arXiv preprint arXiv:1312.7853}, 2013.

\bibitem[Shokri and Shmatikov(2015)]{shokri15privacy}
Reza Shokri and Vitaly Shmatikov.
\newblock Privacy-preserving deep learning.
\newblock In \emph{Proceedings of the 22Nd ACM SIGSAC Conference on Computer
  and Communications Security}, CCS '15, 2015.

\bibitem[Srivastava et~al.(2014)Srivastava, Hinton, Krizhevsky, Sutskever, and
  Salakhutdinov]{srivastava14dropout}
Nitish Srivastava, Geoffrey Hinton, Alex Krizhevsky, Ilya Sutskever, and Ruslan
  Salakhutdinov.
\newblock Dropout: A simple way to prevent neural networks from overfitting.
\newblock 15, 2014.

\bibitem[Sweeney(2000)]{sweeney00simple}
Latanya Sweeney.
\newblock Simple demographics often identify people uniquely.
\newblock 2000.

\bibitem[{TensorFlow team}(2016)]{tensorflowcifar}
{TensorFlow team}.
\newblock Tensorflow convolutional neural networks tutorial, 2016.
\newblock \url{http://www.tensorflow.org/tutorials/deep_cnn}.

\bibitem[{White House Report}(2013)]{whitehouse13privacy}
{White House Report}.
\newblock Consumer data privacy in a networked world: A framework for
  protecting privacy and promoting innovation in the global digital economy.
\newblock \emph{Journal of Privacy and Confidentiality}, 2013.

\bibitem[Yang(2013)]{tianbao13trading}
Tianbao Yang.
\newblock Trading computation for communication: Distributed stochastic dual
  coordinate ascent.
\newblock In \emph{Advances in Neural Information Processing Systems}, 2013.

\bibitem[Zhang and Kwok(2014)]{zhang14admm}
Ruiliang Zhang and James Kwok.
\newblock Asynchronous distributed admm for consensus optimization.
\newblock In \emph{ICML}. JMLR Workshop and Conference Proceedings, 2014.

\bibitem[Zhang et~al.(2015)Zhang, Choromanska, and LeCun]{zhang15elastic}
Sixin Zhang, Anna~E Choromanska, and Yann LeCun.
\newblock Deep learning with elastic averaging sgd.
\newblock In \emph{NIPS}. 2015.

\bibitem[Zhang and Xiao(2015)]{zhang15disco}
Yuchen Zhang and Lin Xiao.
\newblock Communication-efficient distributed optimization of self-concordant
  empirical loss.
\newblock \emph{arXiv preprint arXiv:1501.00263}, 2015.

\bibitem[Zhang et~al.(2012)Zhang, Wainwright, and Duchi]{zhang12communication}
Yuchen Zhang, Martin~J Wainwright, and John~C Duchi.
\newblock Communication-efficient algorithms for statistical optimization.
\newblock In \emph{NIPS}, 2012.

\bibitem[Zhang et~al.(2013)Zhang, Duchi, Jordan, and
  Wainwright]{zhang13information}
Yuchen Zhang, John Duchi, Michael~I Jordan, and Martin~J Wainwright.
\newblock Information-theoretic lower bounds for distributed statistical
  estimation with communication constraints.
\newblock In \emph{Advances in Neural Information Processing Systems}, 2013.

\bibitem[Zinkevich et~al.(2010)Zinkevich, Weimer, Li, and
  Smola]{zinkevich10parallelized}
Martin Zinkevich, Markus Weimer, Lihong Li, and Alex~J. Smola.
\newblock Parallelized stochastic gradient descent.
\newblock In \emph{NIPS}. 2010.

\end{thebibliography}

\clearpage
\appendix

\section{Supplemental Figures and Tables}\label{app:fig}

\begin{figure}[h!]
\begin{center}
\setlength{\pw}{1.6in}
\includegraphics[width=\pw]{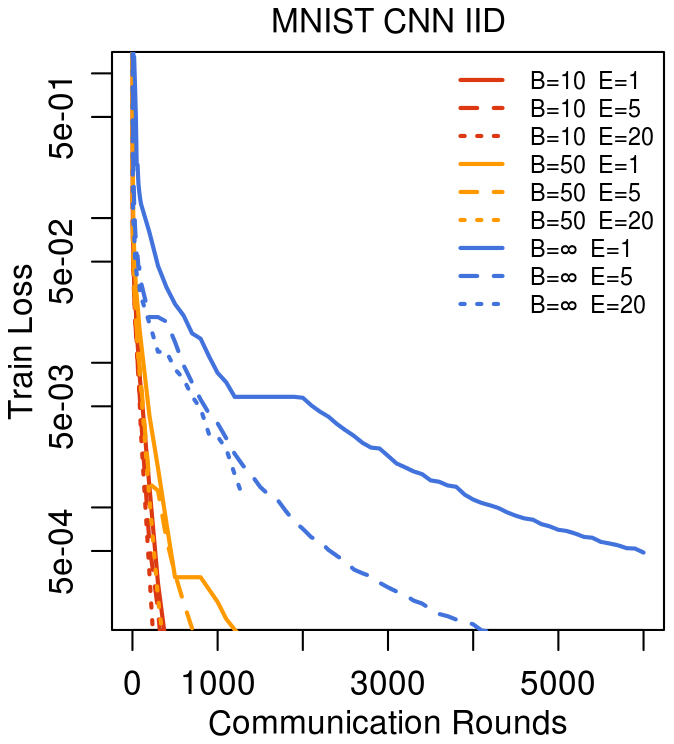}  
\includegraphics[width=\pw]{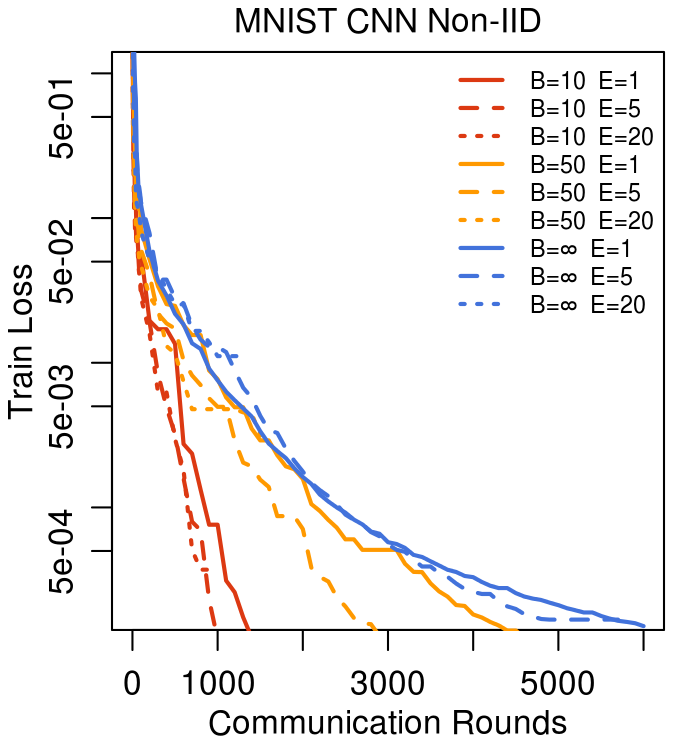}  
\vskip -0.1in
\mycaptionof{figure}{Training set convergence for the MNIST CNN. Note the $y$-axis
  is on a log scale, and the $x$-axis covers more training than 
  Figure~\ref{fig:testaccuracy}. These plots fix $\clientfrac=0.1$.}
\label{fig:mnist-train-loss}
\end{center}
\end{figure}

\begin{figure}[h!]
\setlength{\pw}{1.6in}
\begin{center}
 \includegraphics[width=\pw]{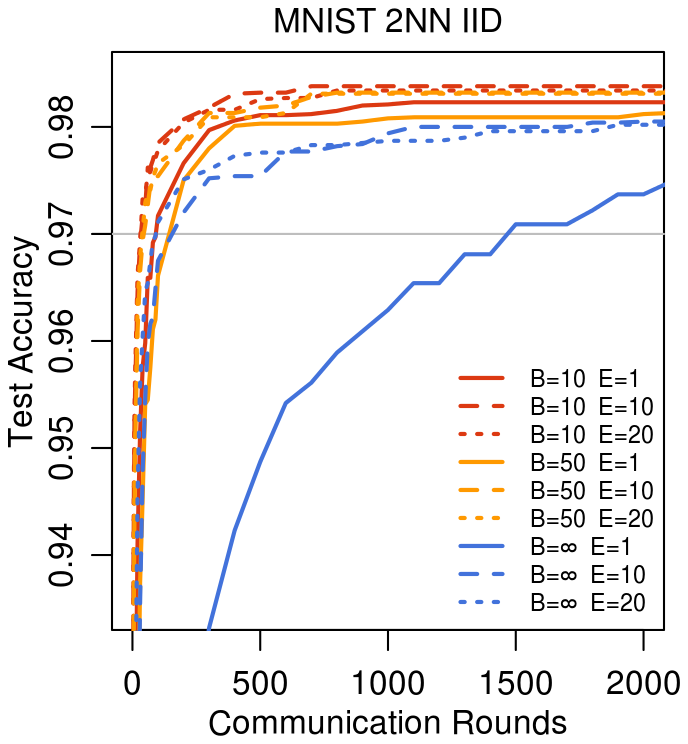}
 \includegraphics[width=\pw]{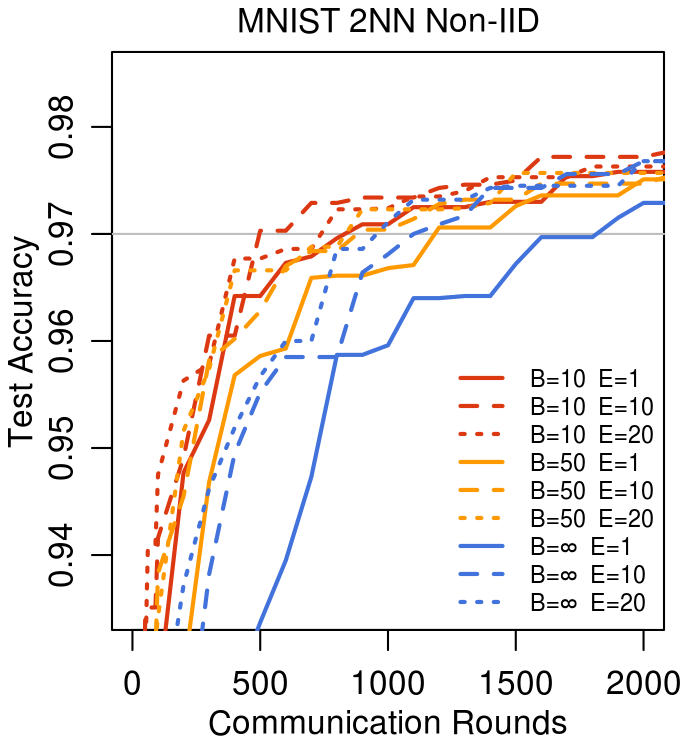}
\mycaptionof{figure}{Test set accuracy vs. communication
  rounds for MNIST \tnn with $\clientfrac=0.1$ and optimized $\eta$. The
  left column is the IID dataset, and right is the pathological
  2-digits-per-client non-IID data. }
\label{fig:mnisttnn}
\end{center}
\end{figure}

\begin{figure}[h!]
\setlength{\pw}{1.6in}
\begin{center}
\includegraphics[width=\pw]{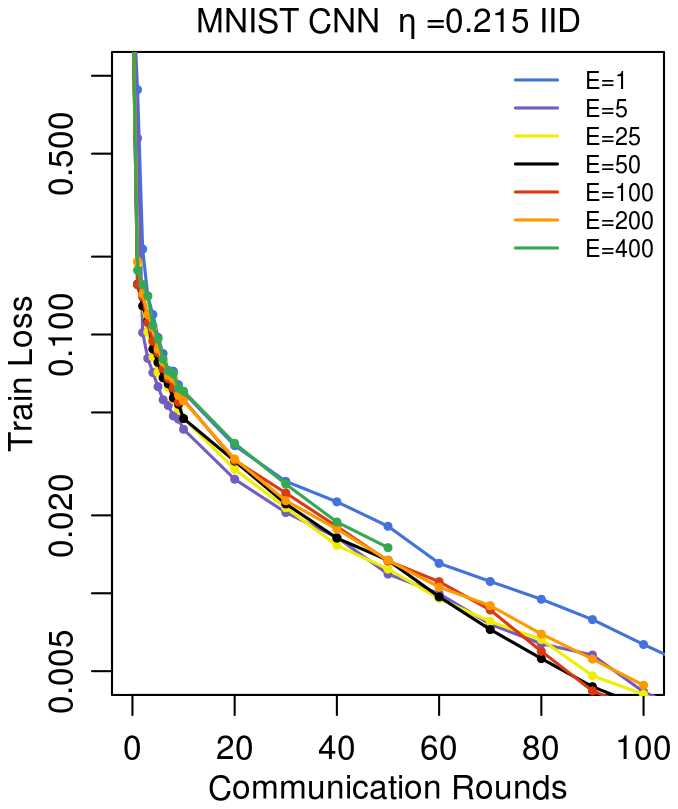}  
\includegraphics[width=\pw]{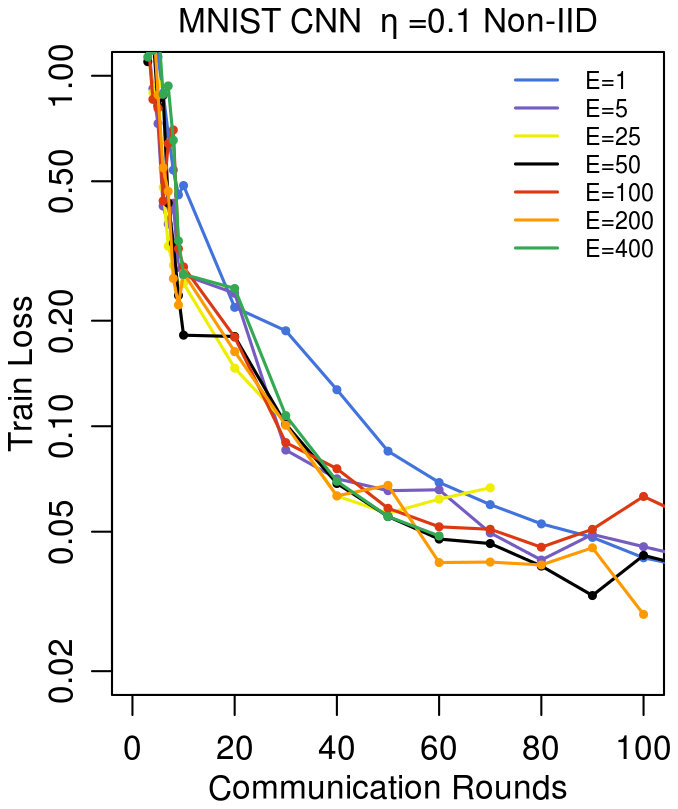}  
\vskip -0.1in 
\mycaptionof{figure}{The effect of training for many
  local epochs (large $\lepochs$) between averaging steps, fixing
  $\lbs=10$ and $\clientfrac=0.1$.  Training loss for the MNIST
  CNN. Note different learning rates and $y$-axis scales are used due
  to the difficulty of our pathological non-IID MNIST
  dataset.}\label{fig:cnn-many-epochs}
  \end{center}
\end{figure}

\begin{table}[h!]
\begin{center}
\mycaptionof{table}{Speedups in the number of communication rounds
    to reach a target accuracy of 97\% for \fedavgshort, versus \fedsgdshort
    (first row) on the MNIST \tnn model.}\label{table:tnn}
\vspace{0.1in}
{\fontsize{7.5pt}{7.5pt}\selectfont
\begin{sc}
\begin{tabular}{rrrrr@{ }rr@{ }r}
\hline
\T \textbf{MNIST \tnn} & $\lepochs$ & $\lbs$ & $u$
 & \multicolumn{2}{c}{IID} 
 & \multicolumn{2}{c}{Non-IID} \\
\hline
\fedsgdshort &    1 &$\all$&    1 &   1468 &             &   1817 &          \T \\
\fedavgshort &   10 &$\all$&   10 &    156 & $(   9.4\xx)$ &   1100 & $(   1.7\xx)$ \\
\fedavgshort &    1 &   50 &   12 &    144 & $(  10.2\xx)$ &   1183 & $(   1.5\xx)$ \\
\fedavgshort &   20 &$\all$&   20 &     92 & $(  16.0\xx)$ &    957 & $(   1.9\xx)$ \\
\fedavgshort &    1 &   10 &   60 &     92 & $(  16.0\xx)$ &    831 & $(   2.2\xx)$ \\
\fedavgshort &   10 &   50 &  120 &     45 & $(  32.6\xx)$ &    881 & $(   2.1\xx)$ \\
\fedavgshort &   20 &   50 &  240 &     39 & $(  37.6\xx)$ &    835 & $(   2.2\xx)$ \\
\fedavgshort &   10 &   10 &  600 &     34 & $(  43.2\xx)$ &    497 & $(   3.7\xx)$ \\
\fedavgshort &   20 &   10 & 1200 &     32 & $(  45.9\xx)$ &    738 & $(   2.5\xx)$ \\
\hline
\end{tabular}
\end{sc}
}
\end{center}
\end{table}

\begin{figure}[h!]
  \begin{center}
    \includegraphics[width=3.2in]{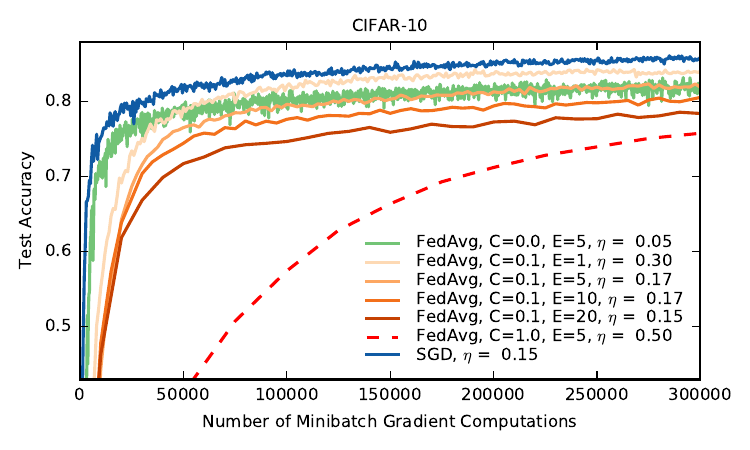} \\
    \vskip -0.05in \mycaptionof{figure}{Test accuracy versus number of
      minibatch gradient computations $(\lbs=50)$. The baseline is
      standard sequential SGD, as compared to \fedavgshort with
      different client fractions $\clientfrac$ (recall $\clientfrac=0$
      means one client per round), and different numbers of local
      epochs $\lepochs$. }\label{fig:cifarbatches}
 \end{center}
\end{figure}

\begin{figure}[h!]
  \begin{center}
    \includegraphics[width=3.2in]{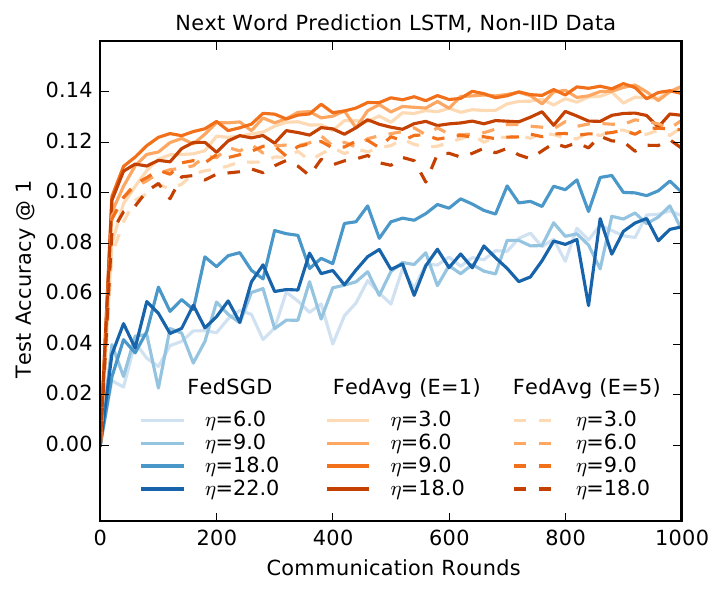} \\
    \vskip -0.05in \mycaptionof{figure}{Learning curves for the
      large-scale language model word LSTM, with evaluation computed
      every 20 rounds. \fedavgshort actually performs better
      with fewer local epochs $\lepochs$ (1 vs 5), and also has lower
      variance in accuracy across evaluation rounds compared to \fedsgdshort.
    }\label{fig:accuracyvar}
 \end{center}
\end{figure}

\end{document}